\documentclass[letterpaper, 10 pt, conference]{ieeeconf}

\usepackage{graphicx}
\usepackage{multirow}
\usepackage{algorithm}
\usepackage{subcaption}
\usepackage{soul}
\usepackage{comment}
\usepackage[noend]{algpseudocode}
\usepackage{afterpage}
\usepackage{array}
\usepackage{amsmath}
\usepackage{amsmath} 
\usepackage{balance}
\usepackage[font=small]{caption}
\usepackage{hyperref}

\IEEEoverridecommandlockouts                              

\begin{document}
\title{\LARGE \bf NMPC-Lander: Nonlinear MPC with Barrier Function for UAV Landing on a Mobile Platform}

\author{Amber Batool, Faryal Batool$^{*}$, Roohan Ahmed Khan$^{*}$, Muhammad Ahsan Mustafa \\Aleksey Fedoseev, and Dzmitry Tsetserukou %
\thanks{The authors are with the Intelligent Space Robotics Laboratory, Center for Digital Engineering, Skolkovo Institute of Science and Technology. 
{\tt \{amber.batool, faryal.batool, roohan.khan, ahsan.mustafa, aleksey.fedoseev, d.tsetserukou\}@skoltech.ru}}
\thanks{*These authors contributed equally to this work.}
}

\maketitle
\begin{abstract}

Quadcopters are versatile aerial robots gaining popularity in numerous critical applications. However, their operational effectiveness is constrained by limited battery life and restricted flight range. To address these challenges, autonomous drone landing on stationary or mobile charging and battery-swapping stations has become an essential capability. In this study, we present NMPC-Lander, a novel control architecture that integrates Nonlinear Model Predictive Control (NMPC) with Control Barrier Functions (CBF) to achieve precise and safe autonomous landing on both static and dynamic platforms. Our approach employs NMPC for accurate trajectory tracking and landing, while simultaneously incorporating CBF to ensure collision avoidance with static obstacles. Experimental evaluations on the real hardware demonstrate high precision in landing scenarios, with an average final position error of 9.0 cm and 11 cm for stationary and mobile platforms, respectively. Notably, NMPC-Lander outperforms the B-spline combined with the A* planning method by nearly threefold in terms of position tracking, underscoring its superior robustness and practical effectiveness. 
Video of NMPC-Lander: \href{https://youtu.be/JAmYipRCiWo}{https://youtu.be/JAmYipRCiWo}
\end{abstract} 
\textbf{\textit{Keywords:}} \textbf{\textit{Non-linear Model Predictive Control, Autonomous Landing, Control Barrier Function, Planning}}


\section{Introduction}
Quadcopters are highly agile aerial robots capable of operating in swarms, and with continuous advancements in drone technology, their ease of deployment and operability have greatly improved. As a result, they are increasingly being adopted in applications such as mapping, surveillance, reconnaissance, search and rescue, infrastructure inspection, and more. Despite this progress, limited battery life and limited operational range remain key challenges that hinder their full potential.
To enhance the autonomy and utility of drones, the concept of autonomous landing on static or moving ground stations primarily for battery charging or swapping has gained significant attention. This capability can greatly benefit mission-critical operations such as search and rescue and drone-based delivery by extending mission time and reducing the need for manual intervention.

In this paper, we propose NMPC-Lander, a novel architecture that leverages Nonlinear Model Predictive Control (NMPC) for drone trajectory generation, control, and precise landing. To ensure safe navigation, we integrate Control Barrier Functions (CBFs) into the control layer of NMPC to avoid obstacles. This combination enables real-time, safe, and efficient landing on both static and dynamic platforms. The key contributions of this paper are:
\begin{itemize}
    \item Development of an NMPC-based controller that observes all 12 states of the drone (position, orientation, linear, and angular velocities), enabling finer control and improved dynamic performance.
    \item Implementation of a single scalable NMPC framework capable of dynamic trajectory generation, path following, target tracking, and landing on both stationary and moving platforms.
    \item Integration of Control Barrier Functions into the NMPC control loop to enforce safety constraints and ensure obstacle avoidance during descent and approach.
    \item Real-world experiments are performed using the proposed NMPC-CBF approach, with all computations performed entirely onboard the drone, demonstrating the practical viability of the system in real-world scenarios.
\end{itemize}

\begin{figure}[t]
\centering
\includegraphics[width=1.0\linewidth]{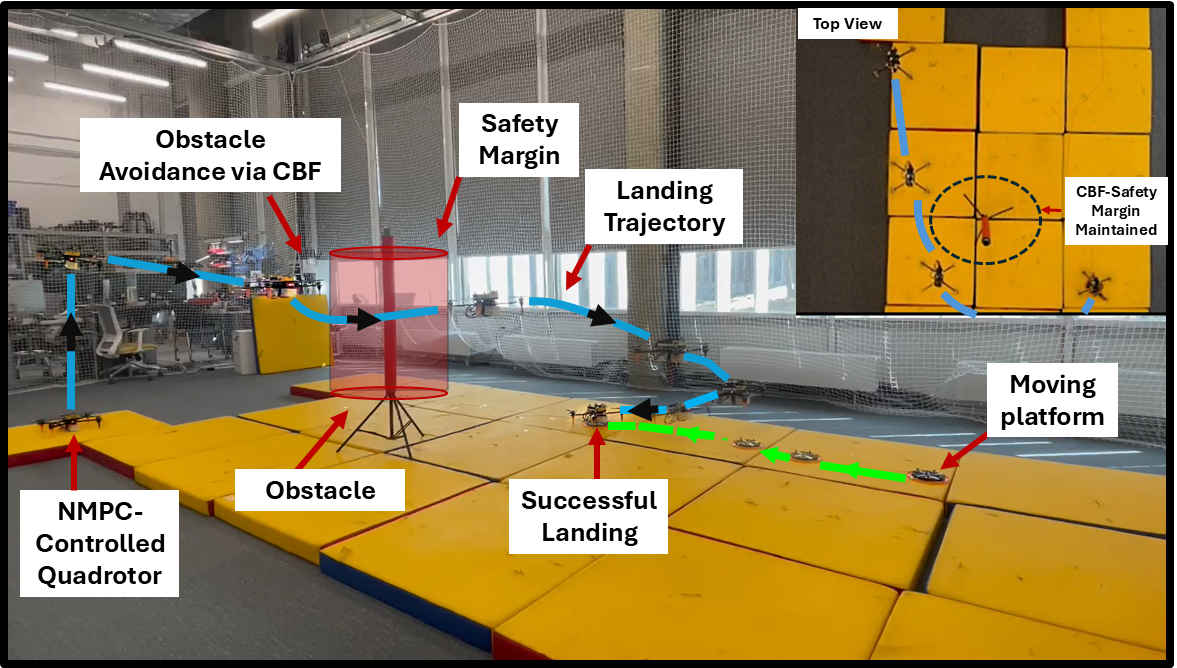}
\label{Title.png}
\caption{NMPC-Lander Technology Demonstration: Drone performing autonomous landing using NMPC-CBF framework}
\vspace{-15pt}
\end{figure}

\section{Related Works}
Autonomous drone landing has remained a challenging and active area of research over the past few decades. A substantial portion of the literature has focused on designing control algorithms or employing learning-based approaches to enable drones to autonomously land on static or moving platforms for battery recharging or swapping. Vision-based systems are widely utilized to detect and track landing platforms due to their robustness and flexibility, although alternative solutions such as GPS-based methods and mock-up target systems have also been explored.

\begin{figure*}[t!]
\centering
\includegraphics[width=1\linewidth]{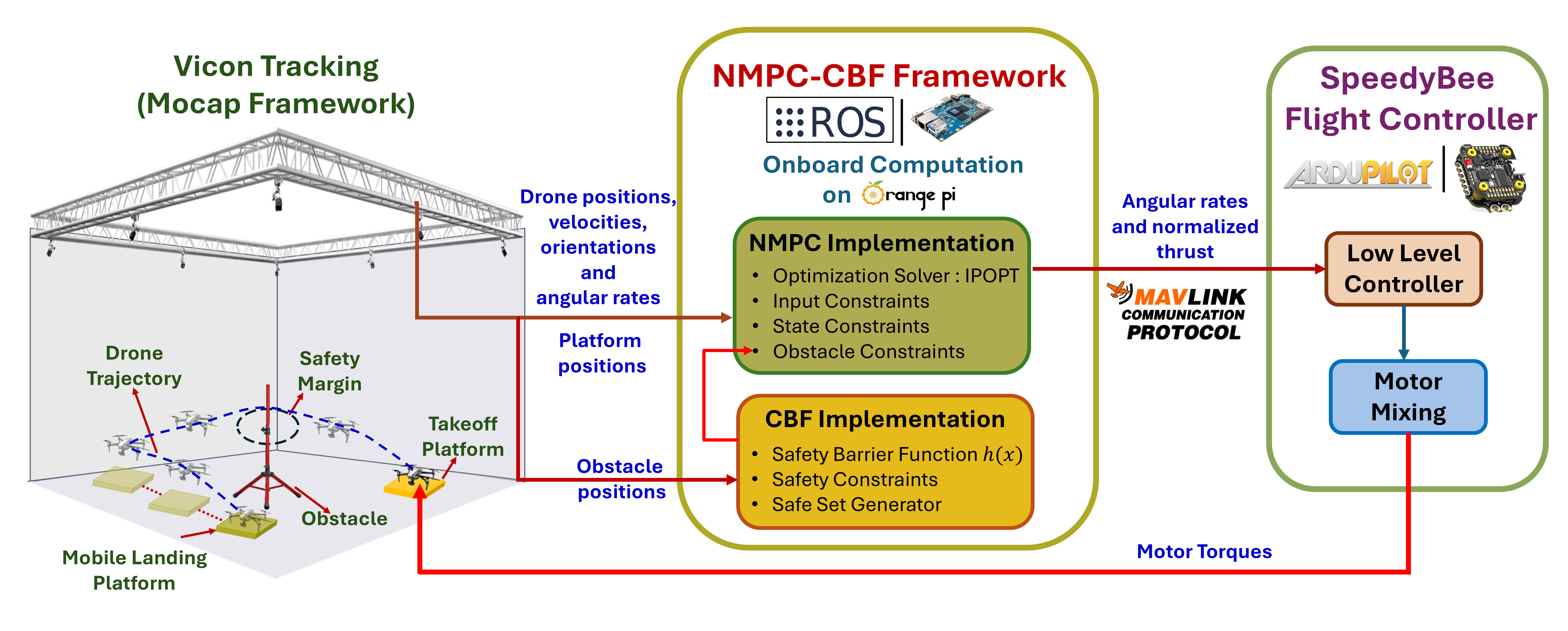}
\caption{System architecture of the NMPC-Lander.}
\label{fig:architecture}
\vspace{-5mm}
\end{figure*}
Drone landing with obstacle avoidance and disturbance rejection constitutes a nonlinear control problem, for which both linear and nonlinear Model Predictive Control (MPC) strategies have been widely used for trajectory planning and tracking.
For example, Pozzan et al. \cite{Paper_01} proposed a hybrid control architecture that combines NMPC for trajectory tracking with PID controllers for stability during landings on moving platforms. Zhu et al. \cite{Paper_02} enhanced NMPC by integrating an Unscented Kalman Filter (UKF) to improve disturbance rejection capabilities. Several other studies, including those by A. et al. \cite{Paper_03}, Mohammadi et al. \cite{Paper_04}, and Feng et al. \cite{Paper_05}, employed vision-based state estimation systems in conjunction with linear MPC for trajectory generation. Paris et al. \cite{Paper_06} and Guo et al. \cite{Paper_07} adopted similar approaches, albeit with varying control strategies tailored to handle aerodynamic disturbances and ensure precise trajectory following. Falanga et al. \cite{Paper_08} introduced minimum-jerk optimization for smooth trajectory planning, complemented by a nonlinear controller and vision-based feedback for accurate landings. However, these approaches primarily addressed disturbances during landing without explicitly incorporating collision avoidance mechanisms. 
Fedoseev et al. \cite{Paper_24} proposed an approach to support the docking process by vehicle.

Research has also expanded to cooperative landing scenarios. Rabelo et al. \cite{Paper_09} and Muskardin et al. \cite{Paper_10} explored drone-UGV collaboration for landing, whereas Aschu et al. \cite{Paper_11}, Peter et al. \cite{Paper_12}, and Gupta et al. \cite{Paper_13} studied swarm-based landing strategies using multi-agent deep reinforcement learning, single-agent deep reinforcement learning, and leader-follower approaches, respectively. Gupta et al. \cite{Paper_14} further introduced artificial potential fields (APFs) to prevent intra-swarm collisions. Although planners like APF effectively avoid obstacles, they generally do not explicitly manage drone oscillations or regulate motion rates.

Obstacle avoidance during landing is critical, especially in densely cluttered environments. Lindqvist et al. \cite{Paper_15} employed NMPC for dynamic obstacle prediction and avoidance. Similarly, Kamel et al. \cite{Paper_16} proposed a unified NMPC framework designed for reactive collision avoidance and trajectory tracking among multiple drones. Guo et al. \cite{Paper_17} used NMPC for control and integrated B-spline with the A* algorithm for path planning in obstructed environments. Recently, deep learning for visual obstacle detection was suggested; e.g., Lee and Kwon \cite{Paper_18} utilized YOLO-based perception to guide motion control during drone descent.

Despite these advances, current obstacle avoidance methods that use artificial intelligence, potential fields, or trajectory replanning often lack formal safety guarantees. In response, Control Barrier Functions (CBFs) have gained popularity, offering rigorous safety strategies for autonomous systems. Tayal et al. \cite{Paper_19} applied CBFs to avoid collisions with kinematic obstacles, while Mundheda et al. \cite{Paper_20} and Folorunsho et al. \cite{Paper_21} integrated CBFs with linear MPC and nominal position controllers for constrained operational environments.

However, the integration of Control Barrier Functions with full-state NMPC for autonomous drone landing remains underexplored. This paper aims to bridge this gap by proposing a unified NMPC-CBF framework, integrating NMPC for optimal control with CBF for formal safety guarantees, within a single architecture designed specifically for real-time onboard deployment.

\section{NMPC-Lander Technology}
\subsection{Quadrotor Model}
In the context of drone dynamics, the full state of the system is represented as a 12-dimensional state vector, combining position, velocity, orientation (Euler angles), and angular velocity. The position of the drone in the world frame is denoted by the vector \( p_w = [p_x, p_y, p_z]^T \), representing the drone's location along the x, y, and z axes. The corresponding linear velocity in the world frame is given by \( v_w = [v_x, v_y, v_z]^T \), which represents the rate of change in position along each axis. The orientation of the drone is typically expressed in terms of Euler angles: roll \( (\phi) \), pitch \( (\theta) \), and yaw \( (\psi) \), which correspond to rotations about the body-fixed x, y, and z axes, respectively. This can be written as the orientation vector \( q_b = [\phi, \theta, \psi]^T \). Finally, the angular velocity of the drone in the body frame is represented by \( \omega_b = [\omega_x, \omega_y, \omega_z]^T \), which corresponds to the rates of change of roll, pitch, and yaw.

To account ground effects, the drone's dynamics are modified by incorporating a ground effect \cite{Ground_effect} factor into the vertical acceleration which is given by:
\begin{equation}
\dot{v}_z = \frac{1}{m} \cdot \left( \cos(\phi) \cdot \cos(\theta) \right) \cdot T_{\text{orig}} \cdot T_{\text{IGE}} - g,
\end{equation}
where
\begin{equation}
T_{\text{IGE}} = T_{\text{orig}} \cdot \left( 1 - \left( \frac{r}{4 \cdot (z_r + \epsilon)} \right)^2 \right)
\end{equation}

Here, \( r \) is the rotor radius, \( z_r \) is the vertical distance from the rotor to the ground, \( \epsilon \) is a small value to avoid division by zero, \( T_{\text{orig}} \) is the thrust generated by the rotor outside of ground effect, and \( T_{\text{IGE}} \) is the thrust generated by the rotor in ground effect, which includes both the normal thrust and the additional thrust due to the ground effect. This modification allows the drone to counteract the increased lift experienced during landing by adjusting the vertical acceleration accordingly.

\begin{figure}[htbp]
\centering
\includegraphics[width=0.8\linewidth]{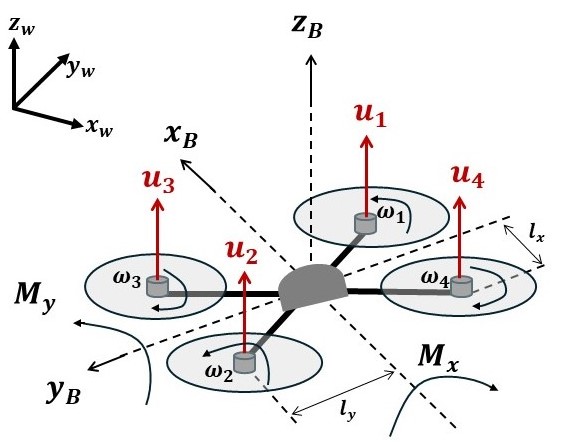}
\caption{Quadrotor Reference System}
\label{Quadrotor1.png}
\end{figure}

The body torque matrix is defined based on the motor configuration (see Fig. \ref{Quadrotor1.png}) as follows:
\[
\begin{bmatrix}
T_x \\
T_y \\
T_z
\end{bmatrix}
=
\begin{bmatrix}
-l_y & l_y & l_y & -l_y \\
-l_x & l_x & -l_x & l_x \\
k_t & k_t & -k_t & -k_t
\end{bmatrix}
\begin{bmatrix}
u_1 \\
u_2 \\
u_3 \\
u_4
\end{bmatrix}
\]
where $T_x$, $T_y$, and $T_z$ are the roll, pitch, and yaw torques, \( l_x \) and \( l_y \) are the distances from the center of mass to the respective control surfaces, \( k_t \) is the constant that scales the thrust based on the characteristics of the motor.
The total thrust $F_{\text{total}}$ is given by:
\begin{equation}
F_{\text{total}} = u_1 + u_2 + u_3 + u_4 
\end{equation}

\subsection{Non-Linear Model Predictive Control (NMPC)}
NMPC is implemented using the multiple-shooting technique \cite{Paper_16} to enable point-to-point autonomous landing of a quadrotor on either a stationary or moving platform. The controller jointly plans and tracks the landing trajectory, ensuring a smooth and precise descent by continuously observing all drone states.

The NMPC serves dual purposes as both a trajectory planner and a follower. At each step, it predicts the future states and computes optimal control inputs over a prediction horizon N = 10 and a sampling time T = 0.1 s. This is accomplished by solving a nonlinear optimization problem using the CasADi framework utilizing the Ipopt solver \cite{Paper_23}. It symbolically defines the system dynamics, objective function, and constraints, automatically computing the derivatives needed for efficient optimization of all states. 

To formulate a discretized nonlinear optimal control problem, the Euler method is applied as follows:
\begin{align}
x(k+1) = x(k) + \delta t \cdot f(x(k), u(k)),
\end{align}
where \( x(k) \) and \( u(k) \) denote the state and control input at time step \( k \), \( \delta t \) is the time increment, \( f(x(k), u(k)) \) describes the system dynamics.
The optimization problem for the NMPC multiple-shooting technique has been formulated as follows:
\begin{align}
l(x_u, u) = \|x_u - x_r\|_Q^2 + \|u\|_R^2 \label{eq:state_cost},
\end{align}
where \( x_u \) is the predicted state, \( x_r \) is the reference state, \( Q \) is the state cost weight matrix, and \( R \) is the control cost weight matrix.

The terminal cost, which penalizes the deviation of the final state from the desired state, is defined as:

\begin{align}
J_{\text{terminal}} = \|x_N - x_{rf}\|_{Q_{\text{terminal}}}^2 \label{eq:terminal_cost},
\end{align}
where \( x_N \) is the state at the final time step, \( x_{rf} \) is the desired final state, and \( Q_{\text{terminal}} \) is the terminal cost weight matrix.

The total cost function \( J(x, u) \) is given as:

\begin{align}
\min_{u}\!J(x, u) = \sum_{k=0}^{N-1} l(x_u(k), u(k)) + J_{\text{terminal}} \label{eq:total_cost}
\end{align}
subject to:
\begin{align*}
x(k+1) &= x(k) + \delta t \cdot f(x(k), u(k)), \\
x_0 &= x_{\text{init}}, \\
u_{\min} &\leq u_k \leq u_{\max}, \quad \forall k \in [0, N-1], \\
x_{\min} &\leq x_k \leq x_{\max}, \quad \forall k \in [0, N]
\end{align*}

\subsection{Platform Tracking}

The planned trajectory is continuously updated in response to the movement of the landing platform. As the drone approaches the stationary or dynamically moving landing platform, the position error between the drone and the platform is further minimized by adding an additional positional cost. This allows the drone to accurately track and land on the platform. An additional positional cost is given by:
\begin{align}
\quad J_{\text{pos}} = \lambda_1 (x_i - x_{\text{f}})^2 + \lambda_2 (y_i - y_{\text{f}})^2 + \lambda_3 (z_i - z_{\text{f}})^2,
\label{eq:positional_cost}
\end{align}
where \( (x_{\text{f}}, y_{\text{f}}, z_{\text{f}}) \) represents the position of the landing platform, and \( (x_i, y_i, z_i) \) denotes the UAV's predicted position at time step \( i \). The parameters \( \lambda_1 \), \( \lambda_2 \), and \( \lambda_3 \) are tuning weights that control the relative importance of position tracking in the \( x \), \( y \), and \( z \) directions, respectively.

\subsection {Integration of CBF with NMPC for Obstacle Avoidance}
To enhance safety during landing, CBF is incorporated into the NMPC framework. This allows integrating the safety constraint into the optimization process, making the drone track its landing trajectory while avoiding static obstacles that may lie in its path as it approaches the moving platform. The CBF ensures that the drone maintains a safe distance from any obstacle, preventing potential collisions.
The CBF is formulated by defining the location of the obstacle at $(x_{\text{obs}}, y_{\text{obs}})$. The total safety distance to be maintained from the obstacle is represented by $r_{\text{safe}}$, which includes an additional user-defined safety margin $r_{\text{safety\_margin}}$ = 30 cm.
The CBF function $h(x, y)$ \cite{paper_22} is defined as:
\begin{equation}
h(x, y) = (x - x_{\text{obs}})^2 + (y - y_{\text{obs}})^2 - r_{\text{safe}}^2,
\end{equation}
where the safe distance is given by:
\begin{equation}
r_{\text{safe}} = r_{\text{obs}} + r_{\text{safety\_margin}}
\end{equation}
The CBF constraint \cite{paper_22} integrated with NMPC constraints is defined as  follows:
\begin{equation}
\Delta h \left( x_{t+k|t}, u_{t+k|t} \right) \geq -\gamma h \left( x_{t+k|t} \right),
\end{equation}
where \( k = 0, \dots, N-1 \) , \( h(x_{t+k|t}) \) is the barrier function at time step \( t+k \), \( \Delta h \) is the change in the barrier function, \( x_{t+k|t} \) is the drone's state at time step \( t+k \), \( u_{t+k|t} \) is the control input at time step \( t+k \) and \( \gamma \) = 0.4 is the tuning parameter for the rate of change of the barrier function.

This ensures that the drone maintains a safe distance from the obstacle at each time step by enforcing the constraint on the barrier function.

\subsection{Gazebo Simulation}
\begin{figure}[htbp]
\centering
\includegraphics[width=0.8\linewidth]{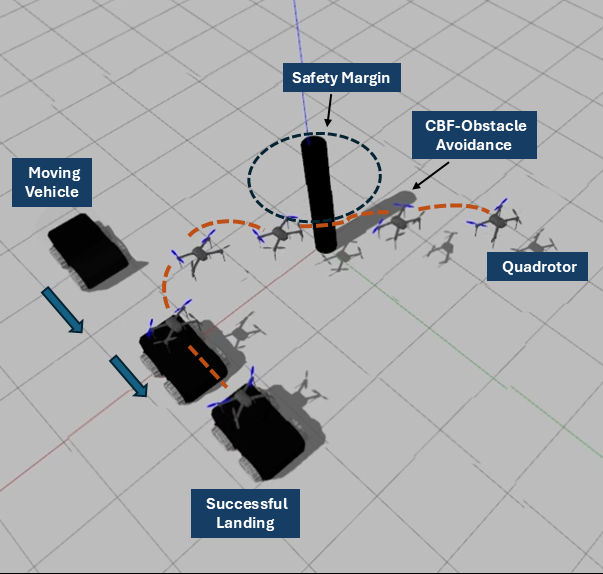}
\caption{Gazebo Simulation for NMPC-CBF Framework}
\label{Gazebo_figure.png}
\end{figure}

After modeling the NMPC controller, it is tuned and tested in the Gazebo simulation environment to evaluate its performance. The simulation setup consists of a quadrotor and a vehicle, where the vehicle serves as the mobile landing platform. As shown in Fig. \ref{Gazebo_figure.png}, the Gazebo environment is integrated with ArduPilot via MAVROS and Robot Operating System (ROS), enabling real-time communication and control.
To assess the effectiveness of integrating Control Barrier Functions (CBFs) with NMPC, static obstacles are placed along the landing path. This setup tests the drone’s ability to safely avoid obstacles while descending.

\section{EXPERIMENTAL EVALUATION}
\subsection{Experimental Setup}
Fig.~\ref{fig:architecture} illustrates the experimental setup used to validate the NMPC-Lander technology. The experiment is conducted on a custom-built quadrotor weighing 1.5 kg. The drone is equipped with an Orange Pi 5B as the onboard computing platform, making it suitable for lightweight, high-performance onboard computation. ROS is used for handling communication, control, and data processing tasks. The controller computes and publishes the angular velocities and normalized thrust using Mavros protocol at 10 Hz. Real-time state feedback is provided by a VICON motion capture system for all the objects in the environment.


\subsection{Results}

\begin{figure}[htbp]
\centering
\includegraphics[width=1.0\linewidth]{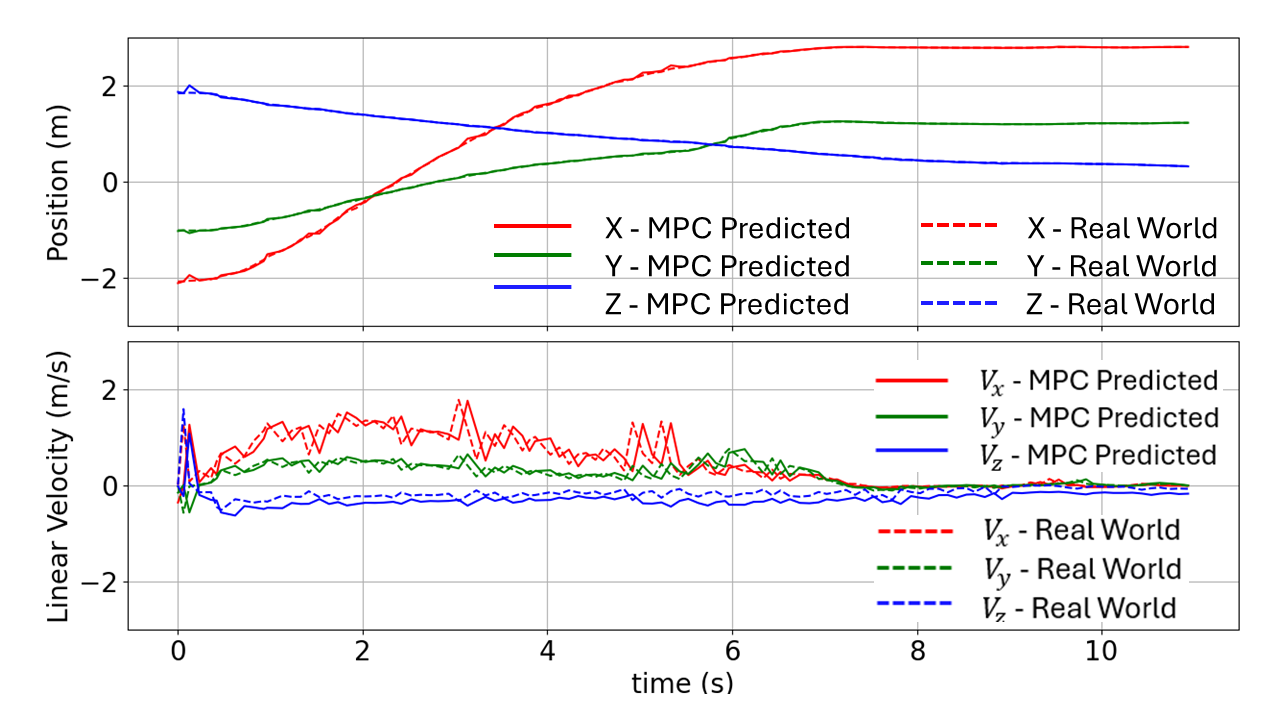}
\caption{Comparison of predicted and observed linear positions and velocities}
\label{pos_velo.png}
\end{figure}

\begin{figure}[htbp]
\centering
\includegraphics[width=1.0\linewidth]{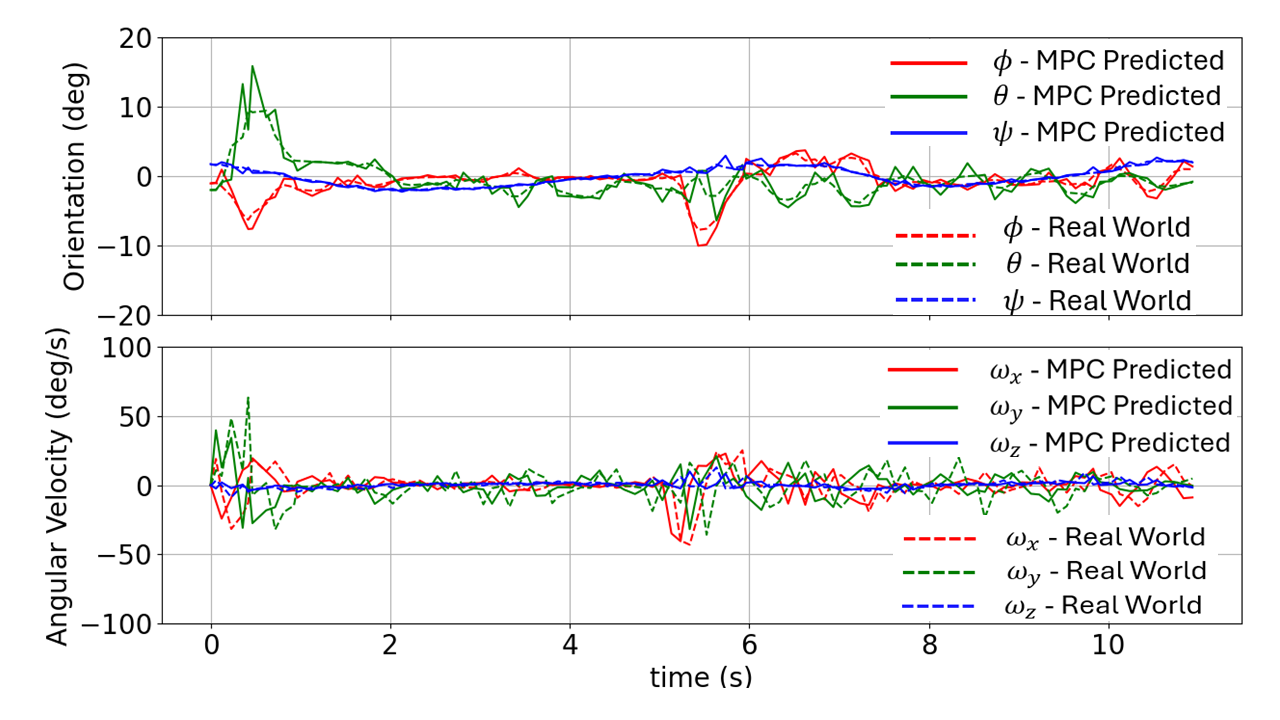}
\caption{Comparison of predicted and observed angular positions and velocities}
\label{oreint_rate.png}
\end{figure}

The proposed NMPC-CBF landing framework was validated across two scenarios: landing on a static platform and a dynamic platform. Each scenario was performed 10 times with and without  obstacles in simulation and real-world environments. Across all the performed experiments, 12 states of the drone were observed, as shown in Fig.~\ref{pos_velo.png} and Fig.~\ref{oreint_rate.png}. It can be seen that NMPC computed predictions align well with real-world observations, showcasing tracking accuracy. 

The evaluation metric used was the average Final Point Error (FPE), defined as the Euclidean distance between the actual and intended landing positions. These errors are shown in Table~\ref{trajectory}.
\\
\begin{table}[h!]
    \centering
    \caption{Final point Error (FPE) comparison between static and dynamic platform landings in both Gazebo simulation and real-world experiments}
    \renewcommand{\arraystretch}{1.3} 
    \begin{tabular}{|>{\centering\arraybackslash}p{2.3cm}|
                    >{\centering\arraybackslash}p{2.3cm}|
                    >{\centering\arraybackslash}p{2.3cm}|}
        \hline
        \textbf{Scenario} & \textbf{Static platform landing FPE (cm)} & \textbf{Dynamic platform landing FPE (cm)} \\
        \hline
        \multicolumn{3}{|c|}{\textbf{Gazebo Simulation}} \\
        \hline
        Without obstacle & 2.1 & 2.5 \\
        \hline
        With obstacle    & 3.4 & 5.2 \\
        \hline
        \multicolumn{3}{|c|}{\textbf{Real World Experiment}} \\
        \hline
        Without obstacle & 3.9 & 6.4 \\
        \hline
        With obstacle    & 9.0 & 11.0 \\
        \hline
    \end{tabular}
    \label{trajectory}
\end{table}

\begin{figure}[htbp]
\centering
\includegraphics[width=1.0\linewidth]{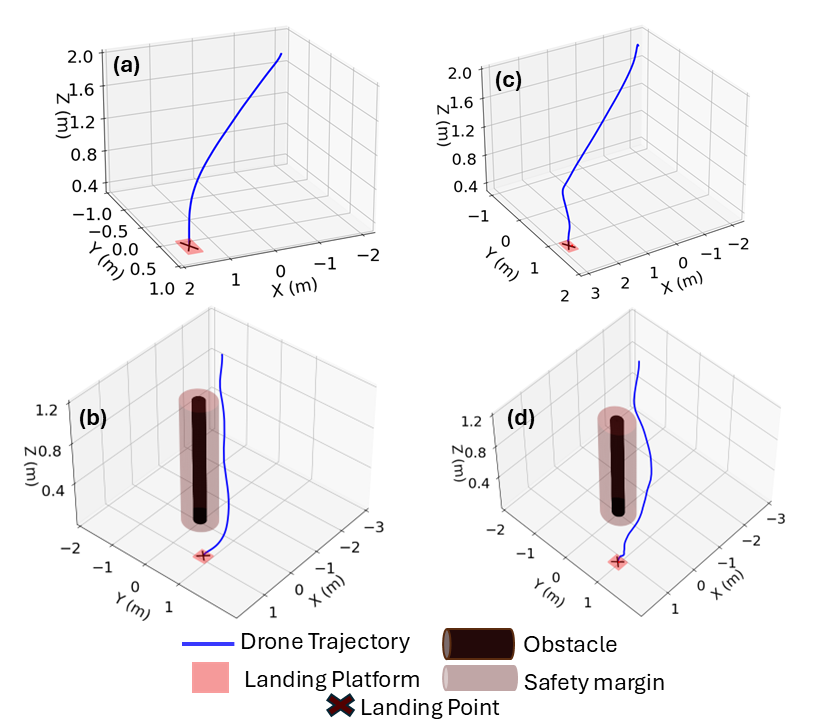}
\label{CASE_1.png}
\caption{Landing visualization: (a)–(b) simulation results in Gazebo, while (c)–(d) real-world landing on a static platform.}
\end{figure}

\begin{figure}[htbp]
\centering
\includegraphics[width=1.0\linewidth]{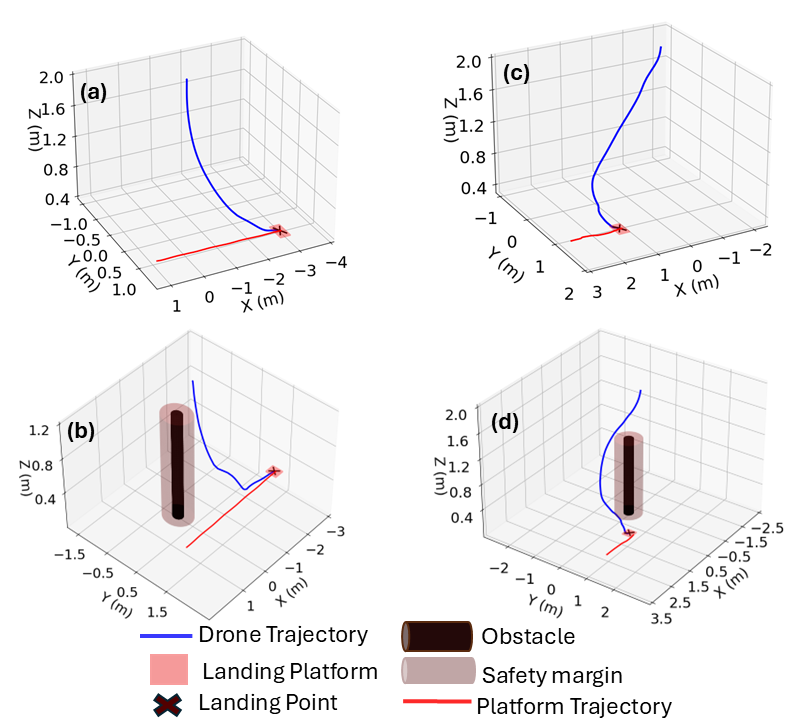}
\label{CASE_2.png}
\caption{Landing visualization: (a)–(b) simulation results in Gazebo, while (c)–(d) real-world landing on a moving platform}
\end{figure}
\subsubsection{Scenario 1: Static Platform Landing}
\subsubsection*{Without Obstacle}
In the absence of obstacles, the NMPC controller achieved highly precise landings. The average final position error was approximately \textbf{2.1--3.9 cm} in both simulation and real-world experiments, demonstrating the controller's baseline accuracy when obstacle constraints are not active.

\subsubsection*{With Obstacle}
Introducing a static obstacle directly in the drone's path required the controller to integrate Control Barrier Function (CBF) constraints for safe avoidance. A minimum safety distance of 30 cm was maintained from the obstacle. The added complexity slightly increased the final point error to approximately \textbf{3.4 cm} and \textbf{9.0 cm} in simulation and real-world settings, respectively.
The average final point error for static platform landings in real-world conditions was approximately 6.5 cm, indicating strong precision and robustness of the NMPC-CBF strategy.

\subsubsection{Scenario 2: Dynamic Platform Landing}

\subsubsection*{Without Obstacle}
For dynamic platforms, the NMPC controller accurately tracked and landed on the target moving with a velocity of 1 m/s. In simulation the FPE was around \textbf{2.5 cm}, whereas in real-world tests where the platform's motion was irregular and manually induced, the FPE increased to \textbf{6.4 cm}. This showcased the controller’s adaptability to real-time disturbances.

\subsubsection*{With Obstacle}
In the presence of an obstacle, the controller leveraged CBF constraints to perform collision avoidance while still tracking the moving platform. This dual-task complexity led to a slight increase in error, with FPE reaching approximately \textbf{11.0 cm} in real-world.

The average final point error for dynamic platform landing in real-world test cases was approximately 8.2 cm, underscoring the reliability of the NMPC-CBF framework in performing simultaneous obstacle avoidance and dynamic target tracking.

\subsection{Comparative Analysis}

To evaluate the effectiveness and precision of our proposed NMPC-Lander framework, we compare our results against B-spline combined with the A* planning algorithm in~\cite{Paper_17} for autonomous UAV landing on moving platforms.

For an equitable comparison, we analyze results under conditions with negligible wind disturbances. Table~\ref{comparison_table} summarizes the final landing point errors reported by both methods.

\begin{table}[h!]
    \centering
    \caption{Comparison of final point error (FPE) between NMPC-Lander and B-spline + A* method under negligible wind conditions}
    \renewcommand{\arraystretch}{1.3}
    \begin{tabular}{|>{\centering\arraybackslash}p{2.5cm}|
                    >{\centering\arraybackslash}p{2.5cm}|
                    >{\centering\arraybackslash}p{2.5cm}|}
        \hline
        \textbf{Approach} & \textbf{Final Point Error (cm)} & \textbf{Platform Velocity (m/s)} \\
        \hline
        B-spline + A* & 34.6 & 0.8 \\
        \hline
        NMPC-Lander & 11 & 1.0 \\
        \hline
    \end{tabular}
    \label{comparison_table}
\end{table}

Despite the higher target velocity of our experiments (1.0 m/s) compared to the B-spline combined with A* method (0.8 m/s), which was evaluated solely in simulation, NMPC-Lander consistently demonstrates lower final point errors across both simulation and real-world scenarios.

In conclusion, the comparative results clearly illustrate the enhanced capability and reliability of the NMPC-Lander system for precise autonomous UAV landing on a mobile platform with obstacle avoidance.

\section{Conclusion and Future Work}
This study presented NMPC-Lander, an integrated NMPC-CBF framework designed to enable precise and safe autonomous landing of quadcopters on both static and moving platforms. By incorporating all drone states into the NMPC algorithm and utilizing CBF for obstacle avoidance, the proposed approach significantly enhances drone autonomy and landing accuracy. Experimental results were confirmed through both Gazebo simulations and rigorous real-world experiments. The system demonstrated robust performance, achieving final position errors consistently within 5.2 cm in simulations and within 11 cm in real-world tests. The experiments effectively validated the system's capability to handle dynamic platform movements and obstacle constraints. Future improvements may focus on optimizing computational efficiency and further reducing landing errors through advanced prediction models and learning strategies. Overall, NMPC-Lander proves to be a reliable and scalable solution for extending the operational capabilities of drones in more complex environments.

\bibliographystyle{IEEEtran}
\balance

\bibliography{BIBILO}
\end{document}